\documentclass[10pt,twocolumn,letterpaper]{article}

\usepackage{cvpr}              
\usepackage{multirow}
\usepackage{amssymb}
\usepackage{pifont}
\newcommand{\cmark}{\ding{51}}  
\newcommand{\xmark}{\ding{55}}  
\usepackage{array}
\usepackage{booktabs}

\definecolor{cvprblue}{rgb}{0.21,0.49,0.74}
\usepackage[pagebackref,breaklinks,colorlinks,allcolors=cvprblue]{hyperref}

\def\paperID{} 
\def\confName{CVPR}
\def\confYear{2026}

\title{UAV-CB: A Complex-Background RGB–T Dataset and Local Frequency Bridge Network for UAV Detection}

\author{
	Shenghui Huang$^{1,2}$ \quad
	Menghao Hu$^{1,4}$ \quad
	Longkun Zou$^{1}$\quad
	Hongyu Chi$^{1,5}$ \quad
	Zekai Li$^{2}$ \quad
	\\Feng Gao$^{3}$ \quad
	Fan Yang$^{3}$ \quad
	Qingyao Wu$^{2}$\thanks{Corresponding author.}\quad
	Ke Chen$^{1}$ \footnotemark[1]\\
	{\small $^{1}$Pengcheng Laboratory, China}\qquad
	{\small $^{2}$South China University of Technology, China}\\
	{\small $^{3}$Peking University, China}\qquad
	{\small $^{4}$Xinjiang University, China}\qquad
	{\small $^{5}$Harbin Institute of Technology, China}\\
	{\small \texttt{ftshhuang@mail.scut.edu.cn; qyw@scut.edu.cn; chenk02@pcl.ac.cn}}
}

\begin{document}
\maketitle
\begin{abstract}
Detecting Unmanned Aerial Vehicles (UAVs) in low-altitude environments is essential for perception and defense systems but remains highly challenging due to complex backgrounds, camouflage, and multimodal interference. In real-world scenarios, UAVs are frequently visually blended with surrounding structures such as buildings, vegetation, and power lines, resulting in low contrast, weak boundaries, and strong confusion with cluttered background textures. Existing UAV detection datasets, though diverse, are not specifically designed to capture these camouflage and complex-background challenges, which limits progress toward robust real-world perception. To fill this gap, we construct UAV-CB, a new RGB–T UAV detection dataset deliberately curated to emphasize complex low-altitude backgrounds and camouflage characteristics. Furthermore, we propose the Local Frequency Bridge Network (LFBNet), which models features in localized frequency space to bridge both the frequency–spatial fusion gap and the cross-modality discrepancy gap in RGB–T fusion. Extensive experiments on UAV-CB and public benchmarks demonstrate that LFBNet achieves state-of-the-art detection performance and strong robustness under camouflaged and cluttered conditions, offering a frequency-aware perspective on multimodal UAV perception in real-world applications.
\end{abstract}    
\section{Introduction}
\label{sec:intro}

\begin{figure}[t]
	\centering
	\includegraphics[width=0.85\linewidth]{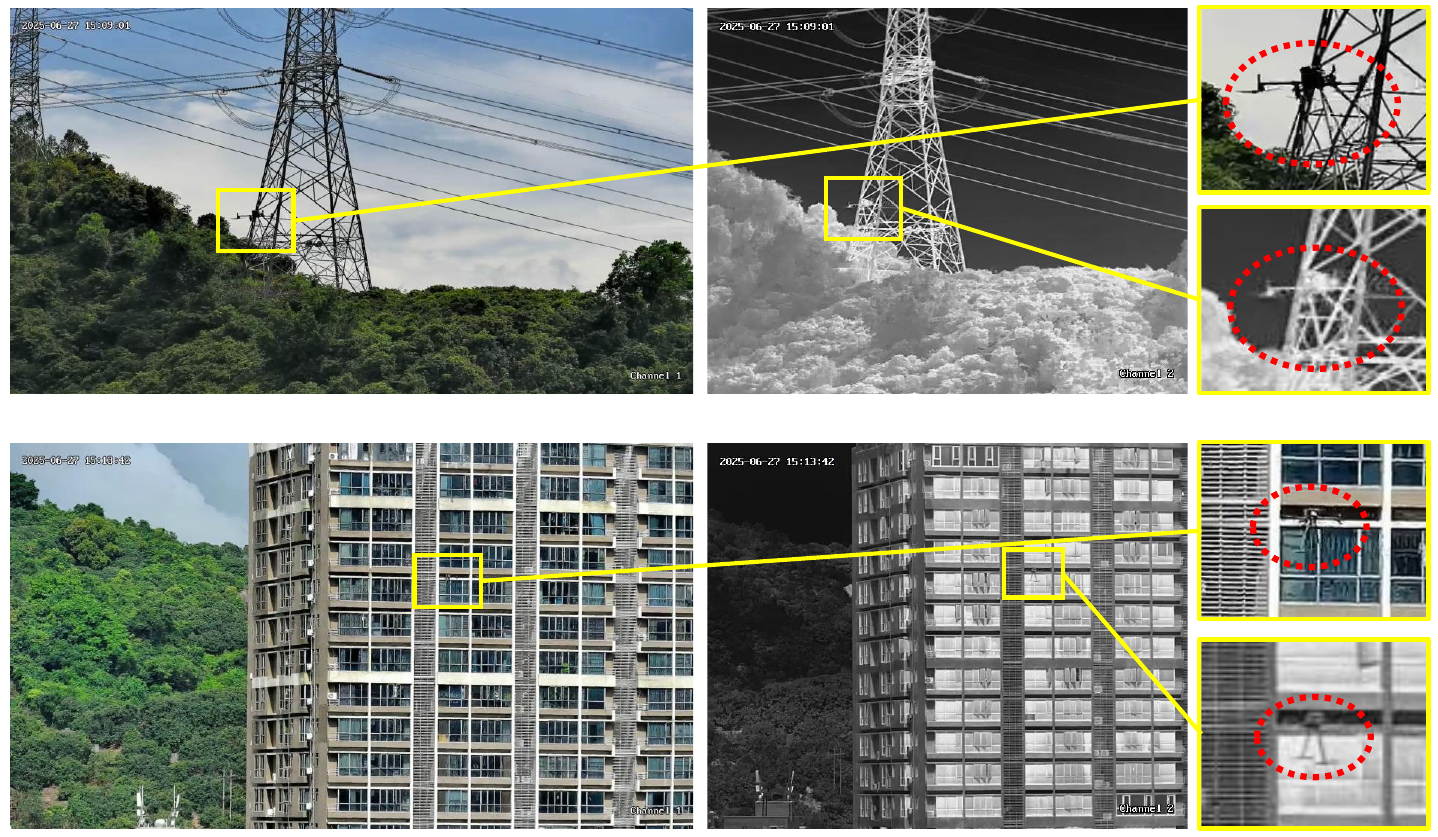}
	\caption{Representative RGB–T examples of UAV detection under complex low-altitude backgrounds.}
	\label{fig:fig1}
\end{figure}

As low-altitude Unmanned Aerial Vehicles (UAVs) become ubiquitous in civilian and defense applications~\cite{c1,c2,c3,c4}, UAV detection is becoming a critical front-end task in perception and defense pipelines. 
It enables subsequent operations such as localization, tracking, identification, and countermeasures~\cite{drones,drones2,UAV_Trajectory_Prediction1,tracking1}.
However, in real-world low-altitude environments, detecting UAVs under complex and cluttered backgrounds remains a persistent challenge.
Background interference often arises from buildings, vegetation, power lines, and cloud layers, forming intricate and dynamic background patterns.
These factors lead to low contrast, weak edges, and strong camouflage for small UAVs~\cite{RCSD-UAV,CST}.
In addition, illumination fluctuations, atmospheric scattering, and motion variations further obscure the distinction between targets and background. 
This severely degrades the robustness of existing vision-based detectors~\cite{Anti-UAV}.
To understand this issue, we examine existing UAV detection benchmarks~\cite{Anti-UAV,AntiUAV410,MMAUD,DUT} and find that, although they cover diverse environments and viewpoints, none are specifically designed to address the challenges posed by camouflage and complex backgrounds.
Most datasets focus on general UAV detection or tracking, without deliberately sampling scenes with strong background interference, environmental clutter, or low-contrast conditions~\cite{Anti-UAV,AntiUAV410}.
Consequently, they offer limited support for analyzing model robustness and failure modes under real-world low-altitude complexity, especially in camouflaged and cluttered backgrounds.
To fill this gap, we construct UAV-CB, an RGB--T UAV detection dataset deliberately designed to capture camouflage and environmental complexity in low-altitude scenes.

Traditional UAV detection methods~\cite{yolo,RCNN} often fail when targets exhibit strong camouflage or appear within cluttered backgrounds.
In such conditions, UAVs usually present low contrast, weak boundaries, and high visual similarity to surrounding textures, leading to both false alarms and missed detections.
While camouflaged object detection (COD)~\cite{COD,COD2} shares a similar notion of visual ambiguity, its formulation is fundamentally different.
COD generally performs pixel-level segmentation on large, static objects in single-modality RGB images.
UAV detection, by contrast, focuses on small, dynamic aerial targets in multimodal low-altitude environments.
Recent studies have explored frequency-domain representations to enhance visual discrimination and suppress interference~\cite{SFDFusion,fd2net}.
However, integrating frequency and spatial representations introduces two critical challenges.
First, there exists a frequency–spatial fusion gap: spatial features capture local structures, while global frequency modeling lacks localized context, resulting in suboptimal cooperation.
Second, there is a cross-modality discrepancy gap, caused by the inherent spectral inconsistency between RGB and thermal modalities, which hampers effective alignment.
To bridge these gaps, we propose the Local Frequency Bridge Network (LFBNet). 
By modeling features in localized frequency space, LFBNet enhances frequency--spatial interactions and leverages the modality-independent nature of frequency representations to achieve better cross-modal alignment.
Furthermore, the proposed local frequency-guided alignment mechanism enables adaptive fusion, leading to robust UAV detection under camouflaged and cluttered backgrounds.
In summary, our main contributions are as follows:
\begin{itemize}
	\item We systematically formalize UAV detection in complex low-altitude environments and highlight the challenges arising from camouflage, low contrast, and multimodal inconsistencies.
	
	\item We introduce UAV-CB, a new RGB--T UAV detection dataset that deliberately focuses on complex backgrounds and camouflage, containing 3,442 RGB--T image pairs across five background categories to support robustness and cross-modality studies.
	
	\item We propose LFBNet, a localized frequency modeling architecture that jointly mitigates the frequency--spatial fusion gap and cross-modality discrepancy, achieving state-of-the-art performance and strong generalization under complex real-world conditions.
\end{itemize}

\section{Related Work}
\label{sec:Related_Work}
\subsection{UAV Detection and Datasets}

UAV detection is crucial for low-altitude security and counter-drone applications. Early studies mainly relied on single-modal detectors such as Faster R-CNN~\cite{fasterrcnn1} and YOLO~\cite{yolo}, focusing on appearance cues from either RGB or thermal imagery~\cite{RCSD-UAV,AntiUAV410}. However, detecting UAVs remains challenging due to their small apparent size, rapid motion, diverse structures, and cluttered real-world backgrounds~\cite{drones}. 
Recent progress has extended UAV perception from single-modal to multimodal sensing, such as RGB–T fusion~\cite{Anti-UAV} and multi-sensor integration~\cite{MMAUD}. However, most existing studies focus on tracking tasks~\cite{Anti-UAV,AntiUAV410}. In practice, detection is the fundamental prerequisite for localization, tracking, and counter-drone operations, yet it remains less explored as an independent problem. Moreover, few efforts explicitly address the complex-background challenge, where small UAVs are easily camouflaged by cluttered scenes. Although recent datasets such as RCSD-UAV~\cite{RCSD-UAV} and CST Anti-UAV~\cite{CST} have begun to consider complex scenarios, they are still single-modal and lack targeted solutions to handle these difficulties. 
To fill this gap, we introduce UAV-CB, a dual-modal RGB–T dataset deliberately curated for complex and camouflaged UAV detection in real-world low-altitude environments.


\begin{table*}[htbp]
	\centering
	\caption{Overview of some representative UAV datasets.}
	\begin{tabular}{p{1.8cm}p{3cm}p{2.2cm}p{8cm}}
		\hline
		\textbf{Modality} & \textbf{Dataset} & \textbf{Data Scale} & \textbf{Description} \\
		\hline
		\multirow{3}{*}{\textbf{Visible}} 
		& Real-World~\cite{Real_World} & 56,821 images & Real-world UAV detection under diverse backgrounds. \\
		& RCSD-UAV~\cite{RCSD-UAV} & 2,844 images & UAV detection in complex real-world scenes. \\
		& DUT-Anti-UAV~\cite{DUT} & 10,000 images & UAV detection and tracking in urban scenarios. \\
		\hline
		\multirow{2}{*}{\textbf{Thermal}} 
		& Anti-UAV410~\cite{AntiUAV410} & 410 sequences & Thermal UAV tracking benchmark for tiny targets. \\
		& CST Anti-UAV~\cite{CST} & 220 sequences & Thermal UAV tracking in complex environments. \\
		\hline
		\multirow{2}{*}{\textbf{Multi-modal}} 
		& Anti-UAV~\cite{Anti-UAV} & 318 sequences & RGB--T UAV detection and tracking benchmark. \\
		& MMAUD~\cite{MMAUD} & 50 sequences & Multi-sensor UAV dataset (RGB, IR, radar, audio). \\
		\hline
	\end{tabular}
\end{table*}

\subsection{RGB--T Fusion}

Research on RGB–T based UAV detection remains relatively limited. Although the Anti-UAV benchmark~\cite{Anti-UAV} introduced an RGB–T dataset for UAV perception, it does not investigate cross-modal detection models or dedicated fusion strategies.
In contrast, RGB–T fusion has been extensively studied in other vision domains such as object detection~\cite{detection,detection2,detection3}, tracking~\cite{tracking,tracking2}, and pedestrian analysis~\cite{pedestrian, pedestrian2}. where complementary visible–thermal cues significantly enhance robustness against illumination variations and background clutter. Existing fusion strategies can be broadly categorized into early, middle, and late fusion~\cite{early-fusion, eml}, employing concatenation, attention mechanisms, or transformer-based cross-modal alignment to integrate modality-specific information~\cite{concatenation,cmx,attention,c2former}. However, most of these designs target generic object scenes, while the challenge of small UAV detection under complex and camouflaged backgrounds remains largely unaddressed.

\subsection{Camouflage-Aware UAV Detection}

Camouflaged object detection (COD) focuses on targets that intentionally blend into backgrounds~\cite{COD,COD2}. Existing COD methods in natural images emphasize boundary refinement, multi-scale aggregation, uncertainty modeling, or contrast priors, and are primarily designed for pixel-level segmentation of large static objects in single-modal RGB~\cite{COD3,COD4,COD5}. 
Adapting COD methods to UAV detection is challenging. UAVs are small, fast-moving targets observed under multimodal low-altitude conditions with dynamic clutter, where localization must remain reliable despite weak edges and low contrast. Nonetheless, the camouflage-aware perspective of COD offers valuable insight, as both tasks involve detecting visually inconspicuous targets in complex backgrounds.
Inspired by this connection, we extend the camouflage concept to multimodal UAV detection and introduce a frequency-based mechanism~\cite{FMNet, SFDFusion} that enhances discriminative cues between UAVs and complex backgrounds.

\section{UAV-CB Dataset}
\label{sec:UAV-CB_Dataset}

\subsection{Data Collection}

\begin{figure}[t]
	\centering
	\includegraphics[width=0.7\linewidth]{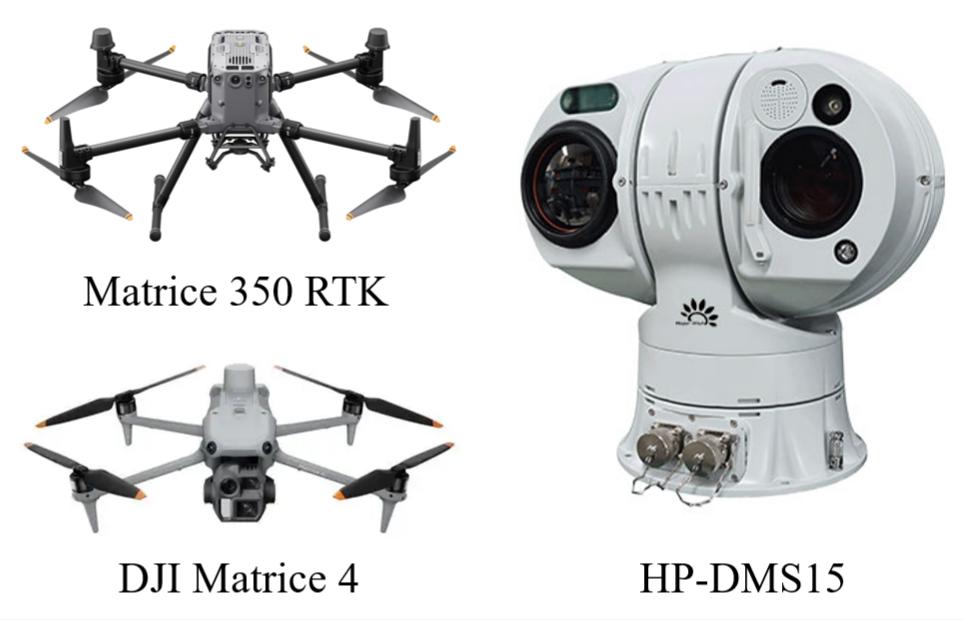}
	\caption{Illustration of the data acquisition platform and the UAV targets used in constructing the UAV-CB dataset.}
	\label{fig:fig2}
\end{figure}

The UAV-CB dataset was captured using an HP-DMS15 electro-optical platform equipped with a co-aligned RGB camera and infrared thermal imager, as Fig.~\ref{fig:fig2}. 
The RGB module adopts a 1/2.8-inch CMOS sensor (1920$\times$1080), while the thermal module employs an uncooled VOx detector (640$\times$512, 8--14\,$\mu$m). 
Both sensors are rigidly mounted and electronically synchronized to ensure precise spatial and temporal alignment across modalities. 
Two UAV models, DJI Matrice~350~RTK (810~$\times$~670~$\times$~430\,mm) and DJI Matrice~4E (307~$\times$~387.5~$\times$~149.5\,mm), 
were used as aerial targets to provide diverse sizes, structures, and reflective characteristics.
To emphasize realistic detection challenges, all samples in UAV-CB were carefully curated from a large corpus of raw RGB--T recordings, 
selecting only those scenes exhibiting substantial background interference and camouflage characteristics. 
The final dataset covers five representative complex background categories--building, vegetation, powerline, cloud, and ground--each introducing distinct spatial structures.

\subsection{Annotation}


During preprocessing, RGB and thermal video streams were synchronized and temporally aligned via hardware timestamps to ensure frame-level correspondence. Annotators manually reviewed the synchronized videos and selected frames with strong camouflage or background confusion to construct the UAV-CB subset. Each selected frame was labeled on both modalities with instance-level bounding boxes—first coarsely on RGB images, then refined on thermal ones for cross-modality consistency. When UAVs were partially occluded or visually blended into the background, annotations were refined using temporal motion cues and thermal evidence. The two modalities were not pixel-level registered, as perfect alignment rarely occurs in real-world systems. This design better reflects practical deployment and encourages research on handling modality misalignment. The resulting dataset provides high-quality annotations suitable for both single-modal and multimodal UAV detection benchmarks.

\subsection{Dataset Statistics and Analysis}
\begin{figure*}[ht]
	\centering
	\includegraphics[width=0.90\linewidth]{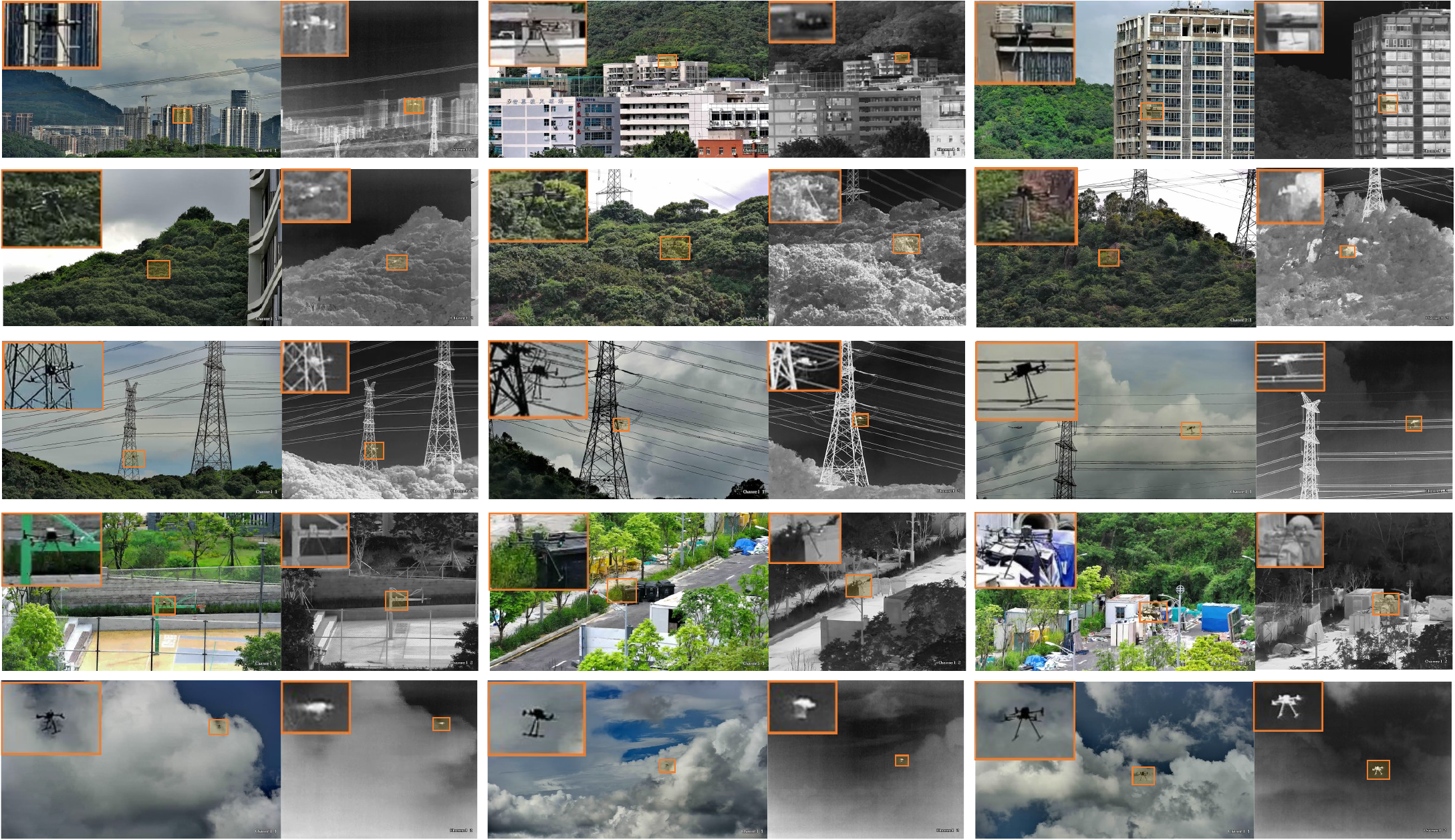}
	\caption{Representative RGB–T image pairs from the UAV-CB dataset covering five complex background categories: buildings, vegetation, power lines, near-ground clutter, and clouds. Each pair shows the visible (left) and thermal (right) modalities with annotated UAV targets (orange boxes). }
	\label{fig:fig3}
\end{figure*}

\begin{figure*}[ht]
	\centering
	\begin{subfigure}[t]{0.28\textwidth}
		\centering
		\includegraphics[width=\linewidth]{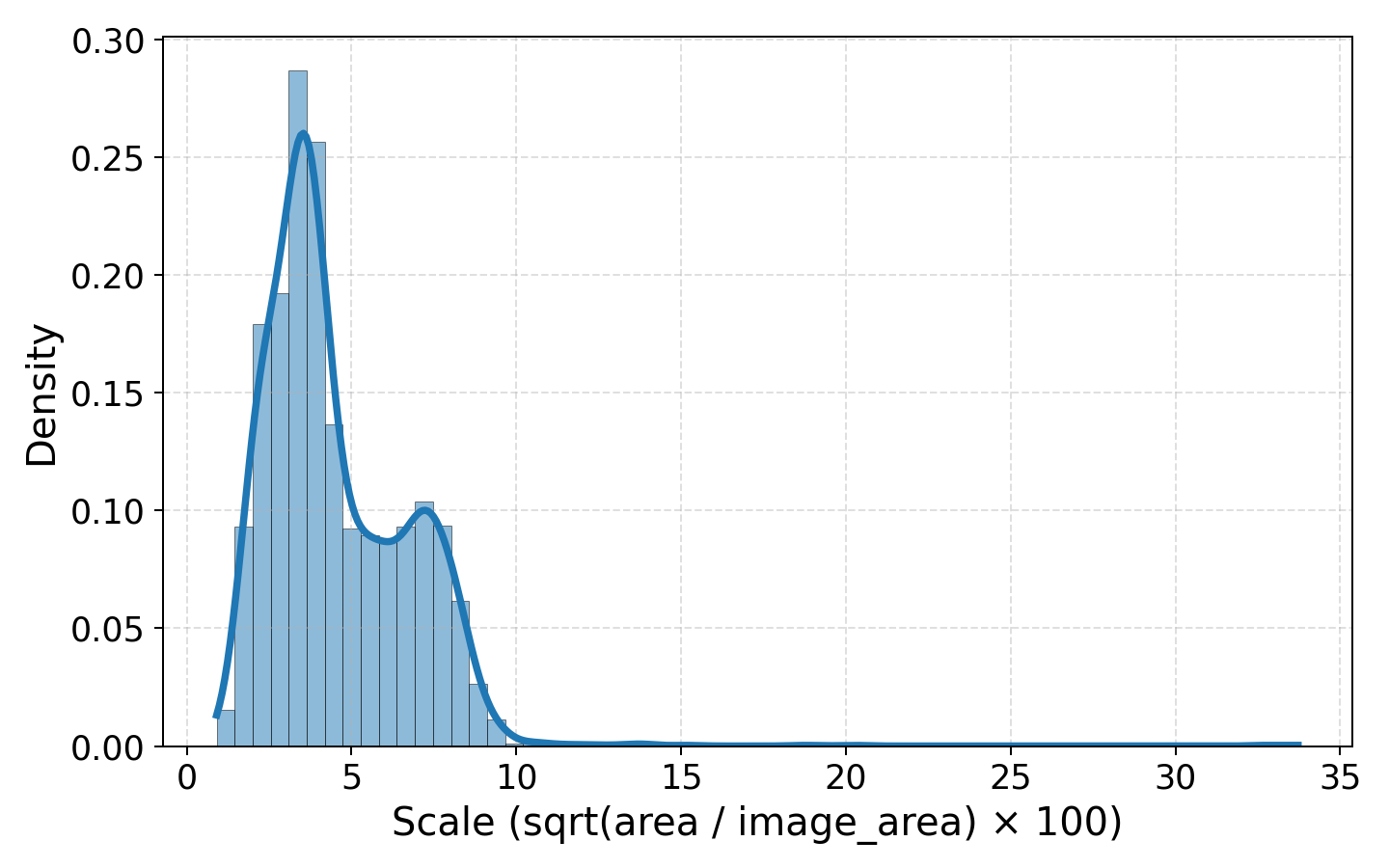}
		\caption{Scale distribution of visible targets.}
		\label{fig:dataset_scale1}
	\end{subfigure}
	\hfill
	\begin{subfigure}[t]{0.28\textwidth}
		\centering
		\includegraphics[width=\linewidth]{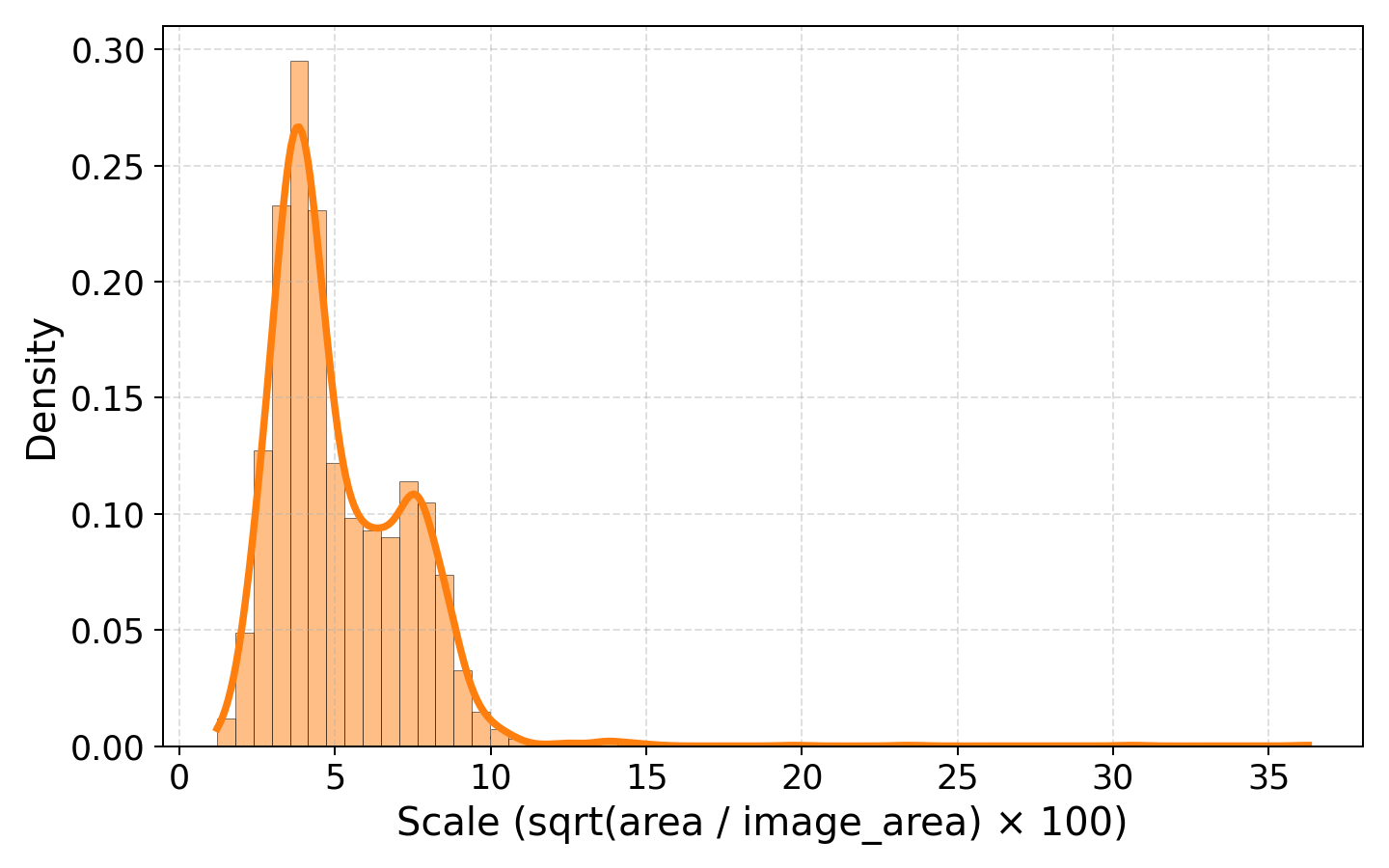}
		\caption{Scale distribution of thermal targets.}
		\label{fig:dataset_scale2}
	\end{subfigure}
	\hfill
	\begin{subfigure}[t]{0.28\textwidth}
		\centering
		\includegraphics[width=\linewidth]{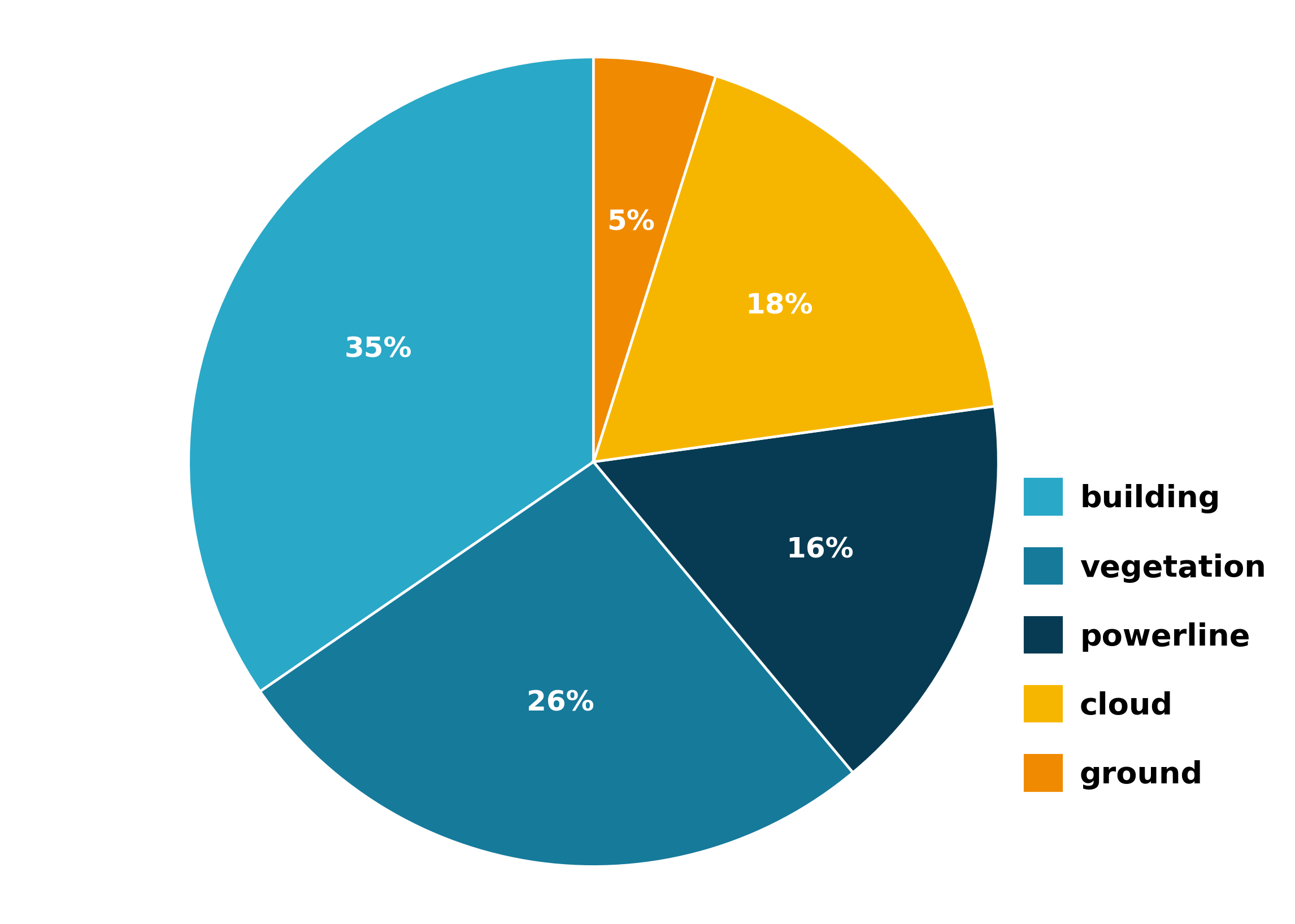}
		\caption{Complex background category distribution.}
		\label{fig:dataset_pie}
	\end{subfigure}
	
	\vspace{-0.5em}
	\caption{Statistical analysis of the proposed UAV-CB dataset.}
	\label{fig:dataset_stats}
\end{figure*}

We conducted a statistical analysis to characterize UAV-CB in terms of scale distribution and background composition.
As illustrated in Fig.~\ref{fig:fig3}, the dataset includes five representative background types—building, vegetation, power line, cloud, and ground—capturing the complexity of real-world low-altitude environments.
As shown in Fig.~\ref{fig:dataset_pie}, samples are relatively balanced across categories, with slightly more instances of building and vegetation scenes, which typically present greater visual complexity.
Fig.~\ref{fig:dataset_scale1} and \ref{fig:dataset_scale2} illustrate the scale distributions of visible and thermal targets, where target scale is defined as the square root of the normalized bounding-box area.
Most UAVs occupy less than 5\% of the image area, underscoring the small-object nature of UAV-CB.
The visible and thermal distributions are closely aligned, indicating accurate temporal correspondence between modalities while preserving appearance discrepancies that make multimodal detection more challenging.

\subsection{Benchmark Tasks and Evaluation Protocols}

To ensure fair comparison and reproducibility, UAV-CB defines a unified RGB–T UAV detection benchmark, where detectors localize UAVs within complex and camouflaged backgrounds using visible and thermal imagery. Each RGB–T pair is annotated with bounding boxes and background category labels, enabling evaluation under both single-modality and multimodal settings.

\textbf{Data split design.} UAV-CB comprises 3,442 carefully curated and meticulously annotated pairs of visible–thermal images. We partition the dataset into training, validation, and test splits at a 6:2:2 ratio, while preserving balanced distributions over background categories, target scales, and UAV types to ensure fair evaluation and minimize bias.

\textbf{Benchmark tasks.} We define two primary settings:
(1) Single-modality detection, where models are trained and evaluated on RGB or thermal images independently to assess the robustness of each modality; and
(2) Multimodal detection, where both modalities are jointly utilized to investigate the effectiveness of cross-modal feature alignment and fusion.

\textbf{Evaluation metrics.} We evaluate detection performance using standard Average Precision (AP) metrics. 
For a predicted box $b_p$ and a ground-truth box $b_g$, the Intersection over Union (IoU) is
\begin{equation}
	\text{IoU}(b_p, b_g) = \frac{|b_p \cap b_g|}{|b_p \cup b_g|}.
\end{equation}
A detection is considered correct if $\text{IoU} \ge \tau$. 
Precision and recall are defined as
\begin{equation}
	\text{Precision} = \frac{TP}{TP + FP}, \qquad 
	\text{Recall} = \frac{TP}{TP + FN}.
\end{equation}
The Average Precision at threshold $\tau$ is computed as the area under the precision–recall curve, denoted as $\text{AP}(\tau)$. 
We report three standard metrics: $\text{AP}_{50}$ ($\tau=0.5$), $\text{AP}_{75}$ ($\tau=0.75$), and the averaged 
\begin{equation}
	\text{AP}_{(0.5:0.95)} = \frac{1}{10}\sum_{\tau=0.5}^{0.95}\text{AP}(\tau),
\end{equation}
which jointly measure coarse and fine localization accuracy, particularly for small and camouflaged UAV targets.

\section{Method}

\begin{figure*}[ht]
	\centering
	\includegraphics[width=0.9\linewidth]{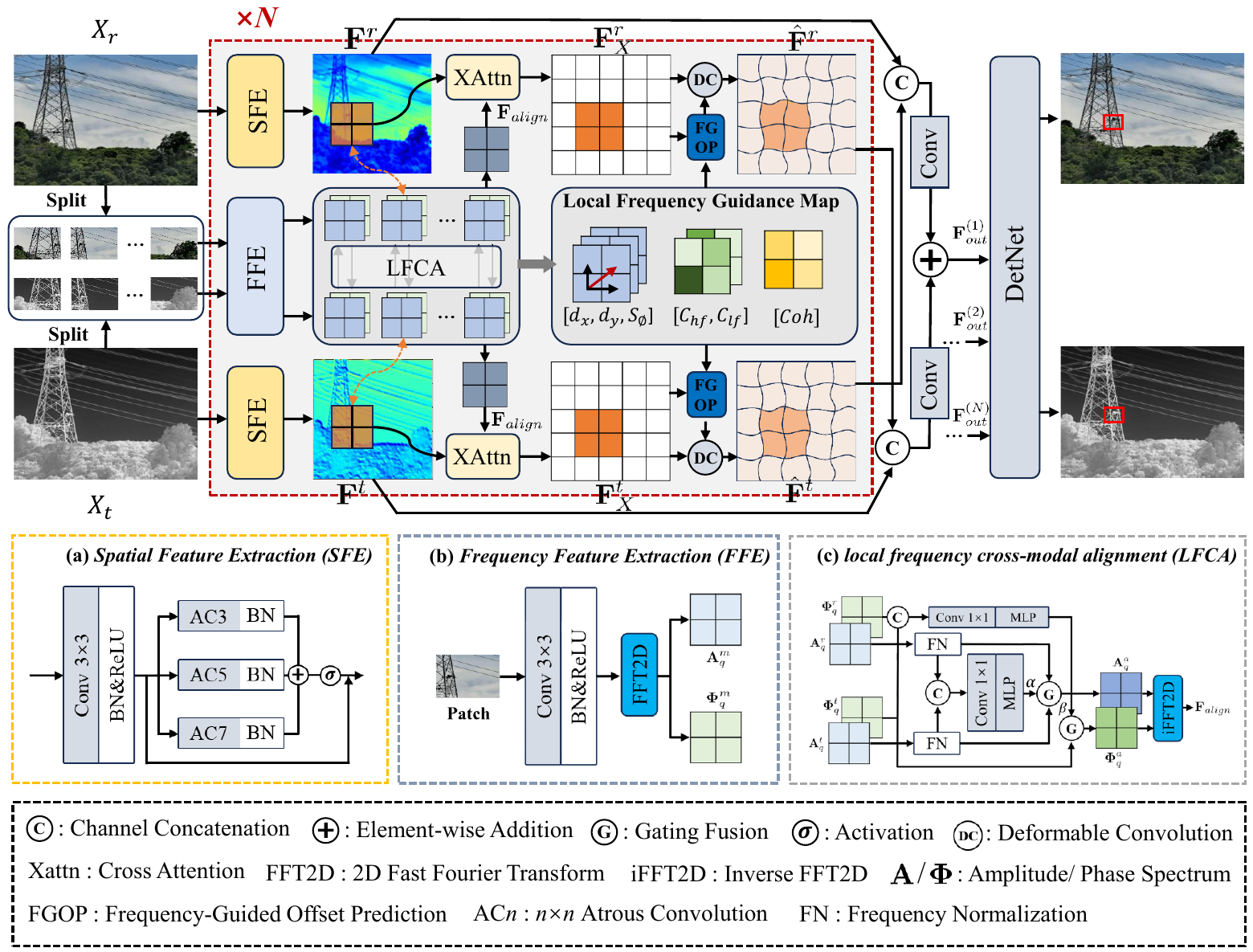}
	\caption{Overview of the proposed LFBNet. The LFCA module aligns RGB and thermal modalities in localized frequency space, while the Local Frequency Guidance Map (LFGM) guides the FGSA module for deformable, frequency-aware spatial fusion. The fused four-scale features ($N=4$) are then fed into an FPN-based detection head (YOLOv5s) for UAV detection.}
	\label{fig:frame}
\end{figure*}

\subsection{Overview}
As illustrated in Fig.~\ref{fig:frame}, LFBNet comprises two main components: Local Frequency Cross-Modal Alignment (LFCA) and Frequency-Guided Spatial Alignment (FGSA). Given paired RGB--T inputs $\mathbf{X}_m$ ($m\!\in\!\{r,t\}$), the network first extracts two complementary representations. Spatial features $\mathbf{F}^m$ are obtained via multi-scale atrous convolutions, while localized frequency features are computed by dividing $\mathbf{X}_m$ into patches $\mathbf{B}^m_q$ and applying a 2D FFT:
\[
\mathcal{F}^m_q = \mathbf{A}^m_q e^{j\boldsymbol{\Phi}^m_q}.
\]
Assembled spectra $\mathcal{F}^m$ provide amplitude--phase cues that complement $\mathbf{F}^m$.

LFCA aligns modalities in localized frequency space and produces a Local Frequency Guidance Map (LFGM). LFGM then guides FGSA to perform frequency-aware deformable fusion in the spatial domain. The fused representation is finally fed into an FPN-based detection head for robust UAV detection under camouflaged and cluttered backgrounds.

\subsection{Local Frequency Cross-Modal Alignment}

Although RGB and thermal modalities differ in illumination and texture, they share consistent geometric structures in the frequency domain: high-frequency components capture edges and shapes, while low-frequency components encode intensity and energy. 
Motivated by this, the Local Frequency Cross-Modal Alignment (LFCA) module aligns the modalities in frequency space through amplitude alignment, phase alignment, and local reconstruction.

\textbf{Amplitude Alignment.}
Given the patch-wise frequency features 
$\{\mathcal{F}^{r}_{q}\}$ and $\{\mathcal{F}^{t}_{q}\}$, each local spectrum can be represented as
\begin{equation}
	\mathcal{F}^{m}_{q} = \mathbf{A}^{m}_{q} e^{j\boldsymbol{\Phi}^{m}_{q}},
	\qquad m \in \{r,t\},
\end{equation}
where $\mathbf{A}^{m}_{q}$ and $\boldsymbol{\Phi}^{m}_{q}$ denote the amplitude and phase spectra of patch $q$ for modality $m$.
Since amplitude reflects the overall energy distribution that varies across modalities, 
we normalize it within each patch to remove scale bias:
\begin{equation}
	\tilde{\mathbf{A}}^{m}_{q} =
	\frac{\mathbf{A}^{m}_{q}}{\|\mathbf{A}^{m}_{q}\|_2 + \epsilon}.
\end{equation}
Then, a learnable coefficient $\alpha_q \in [0,1]$ adaptively balances the normalized amplitudes:
\begin{equation}
	\mathbf{A}^{a}_{q} =
	\alpha_q \tilde{\mathbf{A}}^{r}_{q} + (1-\alpha_q)\tilde{\mathbf{A}}^{t}_{q}.
\end{equation}
This operation harmonizes the spectral energy between modalities, 
ensuring consistent magnitude under varying illumination and emissivity.

\textbf{Phase Alignment.}
The phase spectrum encodes geometric structure. 
To maintain structural consistency, we linearly interpolate between the two modalities’ phases:
\begin{equation}
	\boldsymbol{\Phi}^{a}_{q} =
	\boldsymbol{\Phi}^{t}_{q} +
	\beta_q \cdot \mathrm{wrap}\!\left(\boldsymbol{\Phi}^{r}_{q}-\boldsymbol{\Phi}^{t}_{q}\right),
\end{equation}
where $\mathrm{wrap}(\cdot)$ confines values to $[-\pi,\pi]$, 
and $\beta_q$ controls the contribution of RGB structural details.
This formulation allows thermal geometry to guide coarse alignment, 
while RGB edges refine high-frequency structures.

\textbf{Frequency Reconstruction.}
After amplitude and phase alignment, 
the aligned complex spectrum of each patch is reconstructed as
\begin{equation}
	\mathcal{F}^{a}_{q} = \mathbf{A}^{a}_{q} e^{j\boldsymbol{\Phi}^{a}_{q}}.
\end{equation}
The spatial feature is then obtained via inverse FFT and overlap-add aggregation:
\begin{equation}
	\mathbf{F}_{\mathrm{align}}(x,y)
	= \frac{1}{N_{q(x,y)}}\sum_{q:(x,y)\in q}
	\mathrm{iFFT2D}\!\big(\mathcal{F}^{a}_{q}\big)(x,y),
\end{equation}
where $N_{q(x,y)}$ is the number of overlapping patches covering pixel $(x,y)$.
This yields $\mathbf{F}_{\mathrm{align}}$, a modality-consistent representation.

Finally, cross-attention injects aligned frequency cues into each modality:
\begin{equation}
	\mathbf{F}^{m}_{X} = \mathrm{XAttn}\!\left(\mathbf{F}^{m}, \mathbf{F}_{\mathrm{align}}\right),
	\qquad m \in \{r,t\},
\end{equation}
where $\mathbf{F}^{m}$ denotes the spatial feature from modality $m$.

\subsection{Frequency-Guided Spatial Alignment}

While LFCA achieves cross-modal consistency in the frequency domain, spatial misalignment still exists due to parallax and sensor offset.
To address this, the Frequency-Guided Spatial Alignment (FGSA) module uses spectral cues to guide deformable alignment.

\textbf{Local Frequency Guidance Map (LFGM).} 
To bridge the frequency–spatial domain gap, 
we construct a Local Frequency Guidance Map (LFGM) that converts patch-wise spectral discrepancies 
into spatially interpretable cues. 
Given the aligned frequency patches 
$\{\mathcal{F}^{r}_{q}=\mathbf{A}^{r}_{q} e^{j\boldsymbol{\Phi}^{r}_{q}}\}$ and 
$\{\mathcal{F}^{t}_{q}=\mathbf{A}^{t}_{q} e^{j\boldsymbol{\Phi}^{t}_{q}}\}$ from LFCA, 
LFGM encodes three complementary properties for each patch: 
phase-based displacement, frequency reliability, and spectral coherence.

For each patch $q$, the local phase and energy differences are computed as
\begin{equation}
	\begin{array}{c}
		\Delta\boldsymbol{\Phi}_{q} = \mathrm{wrap}(\boldsymbol{\Phi}^{r}_{q} - \boldsymbol{\Phi}^{t}_{q}), \\[6pt]
		C^{(q)}_{hf} = 
		\dfrac{E^{r,(q)}_{hf}}{E^{r,(q)}_{hf}+E^{t,(q)}_{hf}}, \qquad
		C^{(q)}_{lf} = 
		\dfrac{E^{t,(q)}_{lf}}{E^{r,(q)}_{lf}+E^{t,(q)}_{lf}},
	\end{array}
\end{equation}
where $E^{m,(q)}{hf}$ and $E^{m,(q)}{lf}$ denote the averaged high- and low-frequency energies, 
and $\mathrm{wrap}(\cdot)$ confines phase differences to $[-\pi,\pi]$. 
The local cross-spectral coherence, measuring frequency correlation, is computed as
\begin{equation}
	Coh^{(q)} = 
	\frac{\big|\sum \mathcal{F}^{r}_{q} (\mathcal{F}^{t}_{q})^*\big|}
	{\sqrt{\sum|\mathcal{F}^{r}_{q}|^2\sum|\mathcal{F}^{t}_{q}|^2}}.
\end{equation}

To extract directional and magnitude cues from the phase variation, 
we estimate the local displacement as
\begin{equation}
	\begin{aligned}
		[d_x^{(q)},\, d_y^{(q)}] 
		&= [\sin(\Delta\boldsymbol{\Phi}_{q}),\, \cos(\Delta\boldsymbol{\Phi}_{q})], \\
		S_\phi^{(q)} 
		&= \|\Delta\boldsymbol{\Phi}_{q}\|_1,
	\end{aligned}
\end{equation}
where $(d_x^{(q)},d_y^{(q)})$ indicates the orientation of spectral misalignment and 
$S_\phi^{(q)}$ represents its strength. 
The complete LFGM vector for patch $q$ is defined as
\begin{equation}
	\mathbf{G}^{(q)}_{\mathrm{freq}} = [d_x^{(q)}, d_y^{(q)}, S_\phi^{(q)}, C^{(q)}_{hf}, C^{(q)}_{lf}, Coh^{(q)}].
\end{equation}

To obtain a spatially dense guidance, all patch-wise vectors are projected 
back to the spatial domain via overlap-add interpolation:
\begin{equation}
	\mathbf{G}_{\mathrm{freq}}(x,y) =
	\frac{1}{N_{q(x,y)}} 
	\sum_{q:(x,y)\in q} 
	\mathbf{G}^{(q)}_{\mathrm{freq}},
\end{equation}
where $N_{q(x,y)}$ is the number of overlapping patches covering pixel $(x,y)$. 
This yields a pixel-wise map 
$\mathbf{G}_{\mathrm{freq}}\in\mathbb{R}^{H\times W\times C_g}$ 
aligned with $\mathbf{F}^{r}$ and $\mathbf{F}^{t}$, 
providing frequency-informed cues for offset prediction in FGOP.

\textbf{Frequency-Guided Offset Predictor (FGOP).}
Taking the enhanced spatial features $\mathbf{F}^{r}_{X}$ and $\mathbf{F}^{t}_{X}$, 
together with the projected frequency guidance map $\mathbf{G}_{\mathrm{freq}}$ as input, 
the FGOP predicts an offset field $\Delta\mathbf{p}$ for deformable alignment.
To adaptively integrate frequency priors, we first apply gated modulation:
\begin{equation}
	\tilde{\mathbf{G}} = 
	\sigma\!\left(\mathrm{Conv}_{1\times1}([\mathbf{F}^{r}_{X}, \mathbf{F}^{t}_{X}])\right)
	\odot \mathbf{G}_{\mathrm{freq}},
\end{equation}
where $\sigma(\cdot)$ denotes the sigmoid function.
The offset predictor $f_{\theta}(\cdot)$ then employs two stacked $3\times3$ convolutions 
followed by a $1\times1$ projection to generate the sampling offsets:
\begin{equation}
	\Delta\mathbf{p} = f_{\theta}\!\left([\mathbf{F}^{r}_{X}, \mathbf{F}^{t}_{X}, \tilde{\mathbf{G}}]\right),
\end{equation}
producing $\Delta\mathbf{p} \in \mathbb{R}^{2K_s \times H \times W}$,
where $K_s$ is the number of sampling points in the deformable convolution.
The predicted offsets encode frequency-informed geometric displacements,
enabling precise spatial realignment guided by local spectral cues.

\textbf{Frequency-Guided Deformable Alignment.}
Given the enhanced features $\mathbf{F}^{r}_{X}$ and $\mathbf{F}^{t}_{X}$ 
and the predicted offsets $\Delta\mathbf{p}$ from FGOP, 
deformable convolution is employed to achieve frequency-guided spatial alignment.
For each spatial position $p_{0}$, the aligned feature is computed as
\begin{equation}
	\begin{aligned}
		\hat{\mathbf{F}}^{m}(p_{0})
		&= \sum_{k=1}^{K_s} w_k \cdot 
		\mathcal{S}\!\left(
		\mathbf{F}^{m}_{X},
		\, p_{0} + p_{k} + \Delta p_{k}(p_{0})
		\right), \\
		&\quad m \in \{r,t\},
	\end{aligned}
\end{equation}
where $p_{k}$ denotes the predefined sampling offset, 
$\Delta p_{k}(p_{0})$ is the frequency-guided displacement predicted by FGOP, 
and $\mathcal{S}(\cdot)$ denotes bilinear interpolation for non-integer sampling locations.

Finally, a symmetric cross-conditioned fusion integrates both modalities:
\begin{equation}
	\mathbf{F}_{\mathrm{out}} =
	\operatorname{Conv}_{3\times3}\!\left([\mathbf{F}^{r},\, \hat{\mathbf{F}}^{t}]\right)
	+ 
	\operatorname{Conv}_{3\times3}\!\left([\mathbf{F}^{t},\, \hat{\mathbf{F}}^{r}]\right).
\end{equation}
This symmetric cross-conditioned fusion enables bidirectional interaction between the RGB and thermal branches,
yielding a geometrically aligned and spectrally consistent multimodal representation. 
The fused feature is subsequently fed into an FPN-based detection head to perform final UAV localization and recognition.

\section{Experiments}
\label{sec:exp}
\begin{table*}[t]
	\centering
	\caption{Comparison of UAV detection performance on the UAV-CB dataset.}
	\setlength{\tabcolsep}{3pt}
	\begin{tabular}{lcccccc}
		\hline
		\textbf{Method} & \textbf{Modality} & \textbf{AP$_{50}$(\%)} & \textbf{AP$_{75}$(\%)} & \textbf{AP$_{(0.5:0.95)}$(\%)} & \textbf{Param(M)} & \textbf{FLOPs(G)} \\
		\hline
		Faster R-CNN (NeurIPS'2015)\cite{fasterrcnn1}      & \multirow{9}{*}{Visible} 		& 50.7 & 29.6 & 28.5 & 41.3 & 75.5 \\
		Cascade R-CNN (CVPR'2018)\cite{CascadeRCNN}  &                           			& 63.8 & 33.5 & 34.7 & 69.2 & 103.0 \\
		YOLOv3 (ArXiv'2018)\cite{yolov3}          &                          				& 63.7 & 35.3 & 36.1 & 61.5 & 62.0 \\
		YOLOv5s (Ultralytics'2020)\cite{yolov5}        &                           			& 65.4 & 37.1 & 37.6 & 7.2 & 13.5 \\
		Cascade RPN (NeurIPS'2019)\cite{cascaderpn}    &                           			& 69.3 & 39.4 & 38.7 & 41.9 & 64.2 \\
		YOLOX-S (ArXiv'2021)\cite{yolox}          &                           				& 66.5 & 34.7 & 37.2 & 9.0 & 10.7 \\
		ViTDet-B (ECCV'2022)\cite{vitdet}         &                           				& 70.9 & 42.1 & 40.6 & 86.0 & 317.0 \\
		RT-DETR (CVPR'2024)\cite{RTDETR}         &                           				& 73.4 & 43.7 & 44.2 & 12.0 & 98.6 \\
		YOLOv13S (ArXiv'2025)\cite{yolov13}         &                           			& 67.8 & 38.7 & 39.1 & 9.0 & 16.8 \\
		\cline{1-7}
		Faster R-CNN (NeurIPS'2015)\cite{fasterrcnn1}      & \multirow{9}{*}{Thermal} 		& 69.1 & 41.7 & 37.0 & 41.3 & 75.5 \\
		Cascade R-CNN (CVPR'2018)\cite{CascadeRCNN}  &                           			& 71.0 & 44.0 & 39.1 & 69.2 & 103.0 \\
		YOLOv3 (ArXiv'2018)\cite{yolov3}          &                           				& 77.2 & 48.7 & 42.4 & 61.5 & 62.0 \\
		YOLOv5s (Ultralytics'2020)\cite{yolov5}        &                           			& 78.2 & 48.1 & 41.2 & 7.2 & 13.5 \\
		Cascade RPN (NeurIPS'2019)\cite{cascaderpn}    &                           			& 76.5 & 51.5 & 47.9 & 41.9 & 64.2 \\
		YOLOX-S (ArXiv'2021)\cite{yolox}          &                           				& 77.8 & 52.4 & 48.1 & 9.0 & 10.7 \\
		ViTDet-B (ECCV'2022)\cite{vitdet}         &                           				& 78.3 & 51.8 & 46.6 & 86.0 & 317.0 \\
		RT-DETR (CVPR'2024)\cite{RTDETR}         &                           				& 79.5 & 52.7 & 49.2 & 12.0 & 98.6 \\
		YOLOv13S (ArXiv'2025)\cite{yolov13}         &                           			& 76.8 & 51.9 & 48.6 & 9.0 & 16.8 \\
		\cline{1-7}
		YOLOv5s+Add (Ultralytics'2020)\cite{yolov5}       & \multirow{6}{*}{RGB+T}    		& 72.8 & 45.2 & 38.5 & 13.5  & 26.4 \\
		YOLOv5s+CMX (IEEE TITS'2023)\cite{cmx}     &                           				& 74.3 & 46.5 & 39.8 & 15.2 & 30.4 \\
		UA-CMDet (IEEE TCSVT'2022)\cite{CMDet}        &                          		 	& 76.3 & 47.1 & 46.4 & -- & -- \\
		C$^2$Former (IEEE TITS'2024)\cite{c2former}      &                           		& 79.4 & 53.8 & 49.1 & 100.8 & 324.0 \\
		SFDFusion(ArXiv'2024)\cite{SFDFusion}      &                           				& 79.2 & 51.8 & 49.0 & 21.5 & 58.5 \\
		\textbf{LFBNet (Ours)} &                    								& \textbf{84.6} & \textbf{57.2} & \textbf{54.4} & 30.2  & 65.2 \\
		\hline
	\end{tabular}
	\label{tab:uavcb_sota}
\end{table*}

\begin{table}[h]
	\centering
	\caption{Performance comparison on the DroneVehicle RGB–T detection benchmark.}
	\setlength{\tabcolsep}{10pt}
	\begin{tabular}{lc}
		\hline
		Method  & mAP$_{50}$ \\ \hline
		UA-CMDet (IEEE TCSVT'2022)\cite{CMDet} & 64.0 \\
		TSFADet (ECCV'2022)\cite{TSFADet}   & 70.4 \\
		C$^2$Former (IEEE TITS'2024)\cite{c2former}  & 72.8 \\
		OAFA (CVPR'2024)\cite{OAFA}    & 79.4 \\
		\textbf{LFBNet (ours)}   & \textbf{80.1} \\ \hline
	\end{tabular}
	\label{tab:map50_methods}
\end{table}


\subsection{Experimental Settings}
All experiments on the UAV-CB dataset are conducted in PyTorch on a single NVIDIA A100 (40GB) GPU, using ResNet-50 as the backbone and an input resolution of $640\times512$. Models are trained for 200 epochs using SGD with a momentum of 0.9, a weight decay of $5\times10^{-4}$, an initial learning rate of 0.01, and cosine annealing, with a batch size of 16. Detection performance is reported using $\text{AP}_{50}$, $\text{AP}_{75}$, and $\text{AP}_{(0.5:0.95)}$. To mitigate modality misalignment, RGB images are coarsely registered to the thermal field of view via cropping, followed by resizing to $640\times512$ for consistent multimodal fusion. The patch number of LFBNet is fixed by the $16\times16$ patch size.
For the public RGB–T ground-target detection benchmark DroneVehicle~\cite{CMDet}, we adopt the same settings except that models are trained for 400 epochs and the input size is set to $640\times640$ to ensure fair comparison with recent RGB–T fusion detectors.

\subsection{Results Comparisons}
We compare our proposed LFBNet with state-of-the-art single-modality and multimodal UAV detectors on the UAV-CB dataset. 
As summarized in Tab.~\ref{tab:uavcb_sota}, LFBNet achieves the best performance across all evaluation metrics.
Specifically, it attains an $\text{AP}_{50}$ of 84.6\%, $\text{AP}_{75}$ of 57.2\%, and $\text{AP}_{(0.5:0.95)}$ of 54.4\%, surpassing the previous best multimodal baseline C$^2$Former by 5.3\%. 
Compared with SFDFusion, which also adopts frequency-domain information, LFBNet improves $\text{AP}_{(0.5:0.95)}$ by 5.4\%, indicating that its localized frequency alignment and frequency-guided spatial fusion provide stronger modality-alignment capability and improved robustness to camouflage in UAV-CB.

To further evaluate the generalization ability of our approach beyond UAV-CB, we conduct comparisons on the public DroneVehicle RGB--T ground-target detection benchmark. As shown in Tab.~\ref{tab:map50_methods}, LFBNet achieves the highest mAP$_{50}$ of 80.1\%, surpassing strong multimodal baselines such as OAFA and C$^2$Former. 
Notably, although LFBNet is designed to address the unique challenges of UAV-CB, it still transfers effectively to DroneVehicle, which differs in sensor configuration, viewpoints, and scene distribution.
This cross-dataset improvement shows that the frequency–spatial alignment strategy learned from UAV-CB does not overfit to its specific characteristics. Instead, it captures generalizable RGB–T fusion principles.
It also highlights the value of UAV-CB as a challenging benchmark that drives the development of robust multimodal detectors.

\subsection{Ablation Study on Key Components}
To analyze the contribution of each component in LFBNet, we perform an ablation study on the UAV-CB validation set (Tab.~\ref{tab:ablation}). Starting from the YOLOv5s+Add baseline, incorporating the LFCA module boosts AP$_{(0.5:0.95)}$ from 38.5\% to 47.9\%. This substantial gain confirms that frequency-domain alignment effectively reduces cross-modal amplitude and phase inconsistencies.
Adding the FGSA module alone further improves AP$_{(0.5:0.95)}$ to 49.3\%, demonstrating its ability to correct geometric misalignment via frequency-guided deformable sampling.
When both modules are combined, LFBNet reaches 54.4\% AP$_{(0.5:0.95)}$, showing strong complementarity: LFCA provides modality-consistent spectral features, while FGSA refines spatial correspondence for precise fusion.
These results validate the motivation behind our two-stage alignment design and underscore the value of jointly exploiting frequency cues and spatial deformability for robust RGB--T UAV detection.

\begin{table}[t]
	\centering
	\caption{Ablation study of key components in LFBNet on the UAV-CB validation set.}
	\label{tab:ablation}
	\begin{tabular}{lccc}
		\toprule
		\textbf{Variant} & LFCA & FGSA & AP$_{(0.5:0.95)}$(\%) \\
		\midrule
		YOLOv5s+Add & \xmark & \xmark & 38.5 \\
		+ LFCA only & \cmark & \xmark & 47.9 \\
		+ FGSA only & \xmark & \cmark & 49.3 \\
		\textbf{Full LFBNet} & \cmark & \cmark & \textbf{54.4} \\
		\bottomrule
	\end{tabular}
\end{table}

\section{Conclusion}
We introduced UAV-CB, an RGB–T UAV detection dataset that captures the camouflage and complex backgrounds characteristic of low-altitude environments, and proposed LFBNet, a local frequency–based detector that bridges both frequency–spatial and cross-modality gaps through the LFCA and FGSA modules.
Extensive experiments on UAV-CB and public benchmarks demonstrate that LFBNet achieves strong performance and robustness under cluttered and camouflaged conditions. 
In future work, we will focus on improving the domain adaptability of LFBNet and developing open-environment UAV detection benchmarks to evaluate generalization across unseen weather, lighting, and scene conditions.

\section*{Acknowledgement}
This work was supported by National Natural Science Foundation of China (NSFC) 62272172 and the Fundamental Research Funds for the Central Universities 2025ZYGXZR095. 
This work is supported in part by the Guangdong Pearl River TalentProgram (Introduction of Young Talent) under Grant No. 2019QN01X246.
{
    \small
    \bibliographystyle{ieeenat_fullname}
    \bibliography{main}

@String(CVPR= {IEEE Conf. Comput. Vis. Pattern Recog.})

@String(ICCV= {Int. Conf. Comput. Vis.})

@String(ECCV= {Eur. Conf. Comput. Vis.})

@String(AAAI = {AAAI})

@String(CVPR  = {CVPR})

@String(ICCV  = {ICCV})

@String(ECCV  = {ECCV})

@inproceedings{RCNN,
  title        = {Rich feature hierarchies for accurate object detection and semantic segmentation},
  author       = {Girshick, Ross and Donahue, Jeff and Darrell, Trevor and Malik, Jitendra},
  booktitle    = {Proceedings of the IEEE Conference on Computer Vision and Pattern Recognition (CVPR)},
  pages        = {580--587},
  year         = {2014},
  organization = {IEEE}
}

@inproceedings{yolo,
  title        = {You Only Look Once: Unified, Real-Time Object Detection},
  author       = {Redmon, Joseph and Divvala, Santosh and Girshick, Ross and Farhadi, Ali},
  booktitle    = {Proceedings of the IEEE Conference on Computer Vision and Pattern Recognition (CVPR)},
  pages        = {779--788},
  year         = {2016},
  organization = {IEEE}
}

@Article{drones,
AUTHOR = {Wang, Bingshu and Li, Qiang and Mao, Qianchen and Wang, Jinbao and Chen, C. L. Philip and Shangguan, Aihong and Zhang, Haosu},
TITLE = {A Survey on Vision-Based Anti Unmanned Aerial Vehicles Methods},
JOURNAL = {Drones},
VOLUME = {8},
YEAR = {2024},
NUMBER = {9},
ARTICLE-NUMBER = {518},
URL = {https://www.mdpi.com/2504-446X/8/9/518},
ISSN = {2504-446X},
DOI = {10.3390/drones8090518}
}

@Article{drones2,
AUTHOR = {Huang, Yuzhuo and Qu, Jingyi and Wang, Haoyu and Yang, Jun},
TITLE = {An All-Time Detection Algorithm for UAV Images in Urban Low Altitude},
JOURNAL = {Drones},
VOLUME = {8},
YEAR = {2024},
NUMBER = {7},
ARTICLE-NUMBER = {332},
URL = {https://www.mdpi.com/2504-446X/8/7/332},
ISSN = {2504-446X},
DOI = {10.3390/drones8070332}
}

@ARTICLE{UAV_Trajectory_Prediction1,
  author={Zhang, Jiandong and Shi, Zhuoyong and Zhang, Anli and Yang, Qiming and Shi, Guoqing and Wu, Yong},
  journal={IEEE Transactions on Aerospace and Electronic Systems}, 
  title={UAV Trajectory Prediction Based on Flight State Recognition}, 
  year={2024},
  volume={60},
  number={3},
  pages={2629-2641},
  doi={10.1109/TAES.2023.3303854}}

@inproceedings{tracking1,
  title={A unified transformer based tracker for anti-uav tracking},
  author={Yu, Qianjin and Ma, Yinchao and He, Jianfeng and Yang, Dawei and Zhang, Tianzhu},
  booktitle={Proceedings of the IEEE/CVF Conference on Computer Vision and Pattern Recognition},
  pages={3036--3046},
  year={2023}
}

@article{c1,
  title={Toward Low-Altitude Airspace Management and UAV Operations: Requirements, Architecture and Enabling Technologies},
  author={Luo, Guiyang and Li, Jinglin and Zhang, Qixun and Feng, Zhiyong and Yuan, Quan and Lin, Yijing and Zhang, Hui and Cheng, Nan and Zhang, Ping},
  journal={arXiv preprint arXiv:2506.08579},
  year={2025}
}

@ARTICLE{c2,
  author={He, Dongxuan and Yuan, Weijie and Wu, Jun and Liu, Ruiqi},
  journal={IEEE Network}, 
  title={Ubiquitous UAV Communication Enabled Low-Altitude Economy: Applications, Techniques, and 3GPP’s Efforts}, 
  year={2025},
  volume={},
  number={},
  pages={1-1},
  doi={10.1109/MNET.2025.3574922}}

@article{c3,
  title={Drones help drones: A collaborative framework for multi-drone object trajectory prediction and beyond},
  author={Wang, Zhechao and Cheng, Peirui and Chen, Mingxin and Tian, Pengju and Wang, Zhirui and Li, Xinming and Yang, Xue and Sun, Xian},
  journal={Advances in Neural Information Processing Systems},
  volume={37},
  pages={64604--64628},
  year={2024}
}

@ARTICLE{c4,
  author={Wang, Huiying and Wang, Chunping and Fu, Qiang and Zhang, Dongdong and Kou, Renke and Yu, Ying and Song, Jian},
  journal={IEEE Transactions on Geoscience and Remote Sensing}, 
  title={Cross-Modal Oriented Object Detection of UAV Aerial Images Based on Image Feature}, 
  year={2024},
  volume={62},
  number={},
  pages={1-21},
  doi={10.1109/TGRS.2024.3367934}}

@ARTICLE{detection,
  author={Zhu, Yaohui and Sun, Xiaoyu and Wang, Miao and Huang, Hua},
  journal={IEEE Transactions on Intelligent Transportation Systems}, 
  title={Multi-Modal Feature Pyramid Transformer for RGB-Infrared Object Detection}, 
  year={2023},
  volume={24},
  number={9},
  pages={9984-9995},
  doi={10.1109/TITS.2023.3266487}}

@ARTICLE{detection2,
  author={Ying, Xinyi and Xiao, Chao and An, Wei and Li, Ruojing and He, Xu and Li, Boyang and Cao, Xu and Li, Zhaoxu and Wang, Yingqian and Hu, Mingyuan and Xu, Qingyu and Lin, Zaiping and Li, Miao and Zhou, Shilin and Liu, Li and Sheng, Weidong},
  journal={IEEE Transactions on Pattern Analysis and Machine Intelligence}, 
  title={Visible-Thermal Tiny Object Detection: A Benchmark Dataset and Baselines}, 
  year={2025},
  volume={47},
  number={7},
  pages={6088-6096},
  doi={10.1109/TPAMI.2025.3544621}}

@inproceedings{detection3,
  author       = {Zhao, Tianyi and Liu, Boyang and Gao, Yanglei and Sun, Yiming and Yuan, Maoxun and Wei, Xingxing},
  title        = {Rethinking Multi-modal Object Detection from the Perspective of Mono-Modality Feature Learning},
  booktitle    = {Proceedings of the IEEE/CVF International Conference on Computer Vision (ICCV)},
  year         = {2025},
  pages        = {6364--6373},
  url          = {https://openaccess.thecvf.com/content/ICCV2025/html/Zhao_Rethinking_Multi-modal_Object_Detection_from_the_Perspective_of_Mono-Modality_Feature_ICCV_2025_paper.html}
}

@ARTICLE{tracking,
  author={Zhang, Haiping and Yuan, Di and Shu, Xiu and Li, Zhihui and Liu, Qiao and Chang, Xiaojun and He, Zhenyu and Shi, Guangming},
  journal={IEEE Transactions on Instrumentation and Measurement}, 
  title={A Comprehensive Review of RGBT Tracking}, 
  year={2024},
  volume={73},
  number={},
  pages={1-23},
  doi={10.1109/TIM.2024.3436098}}

@inproceedings{tracking2,
  title={Visual prompt multi-modal tracking},
  author={Zhu, Jiawen and Lai, Simiao and Chen, Xin and Wang, Dong and Lu, Huchuan},
  booktitle={Proceedings of the IEEE/CVF conference on computer vision and pattern recognition},
  pages={9516--9526},
  year={2023}
}

@inproceedings{pedestrian,
  title={When pedestrian detection meets multi-modal learning: Generalist model and benchmark dataset},
  author={Zhang, Yi and Zeng, Wang and Jin, Sheng and Qian, Chen and Luo, Ping and Liu, Wentao},
  booktitle={European Conference on Computer Vision},
  pages={430--448},
  year={2024},
  organization={Springer}
}

@inproceedings{pedestrian2,
  title={Causal mode multiplexer: A novel framework for unbiased multispectral pedestrian detection},
  author={Kim, Taeheon and Shin, Sebin and Yu, Youngjoon and Kim, Hak Gu and Ro, Yong Man},
  booktitle={Proceedings of the IEEE/CVF Conference on Computer Vision and Pattern Recognition},
  pages={26784--26793},
  year={2024}
}

@INPROCEEDINGS{eml,
  author={Hwang, Soonmin and Park, Jaesik and Kim, Namil and Choi, Yukyung and Kweon, In So},
  booktitle={2015 IEEE Conference on Computer Vision and Pattern Recognition (CVPR)}, 
  title={Multispectral pedestrian detection: Benchmark dataset and baseline}, 
  year={2015},
  volume={},
  number={},
  pages={1037-1045},
  doi={10.1109/CVPR.2015.7298706}}

@ARTICLE{early-fusion,
  author={Zhang, Xue and Cao, Si-Yuan and Wang, Fang and Zhang, Runmin and Wu, Zhe and Zhang, Xiaohan and Bai, Xiaokai and Shen, Hui-Liang},
  journal={IEEE Transactions on Intelligent Vehicles}, 
  title={Rethinking Early-Fusion Strategies for Improved Multispectral Object Detection}, 
  year={2025},
  volume={10},
  number={6},
  pages={3728-3742},
  doi={10.1109/TIV.2024.3462488}}

@article{concatenation,
  title={Multispectral deep neural networks for pedestrian detection},
  author={Liu, Jingjing and Zhang, Shaoting and Wang, Shu and Metaxas, Dimitris N},
  journal={arXiv preprint arXiv:1611.02644},
  year={2016}
}

@ARTICLE{attention,
  author={Zhang, Tianlu and Liu, Xueru and Zhang, Qiang and Han, Jungong},
  journal={IEEE Transactions on Circuits and Systems for Video Technology}, 
  title={SiamCDA: Complementarity- and Distractor-Aware RGB-T Tracking Based on Siamese Network}, 
  year={2022},
  volume={32},
  number={3},
  pages={1403-1417},
  doi={10.1109/TCSVT.2021.3072207}}

@ARTICLE{c2former,
  author={Yuan, Maoxun and Wei, Xingxing},
  journal={IEEE Transactions on Geoscience and Remote Sensing}, 
  title={C²Former: Calibrated and Complementary Transformer for RGB-Infrared Object Detection}, 
  year={2024},
  volume={62},
  number={},
  pages={1-12},
  doi={10.1109/TGRS.2024.3376819}}

@ARTICLE{cmx,
  author={Zhang, Jiaming and Liu, Huayao and Yang, Kailun and Hu, Xinxin and Liu, Ruiping and Stiefelhagen, Rainer},
  journal={IEEE Transactions on Intelligent Transportation Systems}, 
  title={CMX: Cross-Modal Fusion for RGB-X Semantic Segmentation With Transformers}, 
  year={2023},
  volume={24},
  number={12},
  pages={14679-14694},
  doi={10.1109/TITS.2023.3300537}}

@inproceedings{COD,
  title={Camouflaged object detection},
  author={Fan, Deng-Ping and Ji, Ge-Peng and Sun, Guolei and Cheng, Ming-Ming and Shen, Jianbing and Shao, Ling},
  booktitle={Proceedings of the IEEE/CVF conference on computer vision and pattern recognition},
  pages={2777--2787},
  year={2020}
}

@article{COD2,
  title={Decoupling and integration network for camouflaged object detection},
  author={Zhou, Xiaofei and Wu, Zhicong and Cong, Runmin},
  journal={IEEE Transactions on Multimedia},
  volume={26},
  pages={7114--7129},
  year={2024},
  publisher={IEEE}
}

@article{COD3,
  title={Zoomnext: A unified collaborative pyramid network for camouflaged object detection},
  author={Pang, Youwei and Zhao, Xiaoqi and Xiang, Tian-Zhu and Zhang, Lihe and Lu, Huchuan},
  journal={IEEE transactions on pattern analysis and machine intelligence},
  volume={46},
  number={12},
  pages={9205--9220},
  year={2024},
  publisher={IEEE}
}

@article{COD4,
  title={Referring camouflaged object detection},
  author={Zhang, Xuying and Yin, Bowen and Lin, Zheng and Hou, Qibin and Fan, Deng-Ping and Cheng, Ming-Ming},
  journal={IEEE Transactions on Pattern Analysis and Machine Intelligence},
  year={2025},
  publisher={IEEE}
}

@inproceedings{COD5,
  title={Mm-camobj: A comprehensive multimodal dataset for camouflaged object scenarios},
  author={Ruan, Jiacheng and Yuan, Wenzhen and Lin, Zehao and Liao, Ning and Li, Zhiyu and Xiong, Feiyu and Liu, Ting and Fu, Yuzhuo},
  booktitle={Proceedings of the AAAI Conference on Artificial Intelligence},
  volume={39},
  number={7},
  pages={6740--6748},
  year={2025}
}

@article{FMNet,
  title={FMNet: Frequency-Assisted Mamba-Like Linear Attention Network for Camouflaged Object Detection},
  author={Deng, Ming and Sun, Sijin and Li, Zihao and Hu, Xiaochuan and Wu, Xing},
  journal={arXiv preprint arXiv:2503.11030},
  year={2025}
}

@article{SFDFusion,
  title={SFDFusion: an efficient spatial-frequency domain fusion network for infrared and visible image fusion},
  author={Hu, Kun and Zhang, Qingle and Yuan, Maoxun and Zhang, Yitian},
  journal={arXiv preprint arXiv:2410.22837},
  year={2024}
}

@ARTICLE{DUT,
  author={Zhao, Jie and Zhang, Jingshu and Li, Dongdong and Wang, Dong},
  journal={IEEE Transactions on Intelligent Transportation Systems}, 
  title={Vision-Based Anti-UAV Detection and Tracking}, 
  year={2022},
  volume={23},
  number={12},
  pages={25323-25334},
  doi={10.1109/TITS.2022.3177627}}

@article{Anti-UAV,
  title={Anti-UAV: A large multi-modal benchmark for UAV tracking},
  author={Jiang, Nan and Wang, Kuiran and Peng, Xiaoke and Yu, Xuehui and Wang, Qiang and Xing, Junliang and Li, Guorong and Zhao, Jian and Guo, Guodong and Han, Zhenjun},
  journal={arXiv preprint arXiv:2101.08466},
  year={2021}
}

@ARTICLE{AntiUAV410,
  author={Huang, Bo and Li, Jianan and Chen, Junjie and Wang, Gang and Zhao, Jian and Xu, Tingfa},
  journal={IEEE Transactions on Pattern Analysis and Machine Intelligence}, 
  title={Anti-UAV410: A Thermal Infrared Benchmark and Customized Scheme for Tracking Drones in the Wild}, 
  year={2024},
  volume={46},
  number={5},
  pages={2852-2865},
  doi={10.1109/TPAMI.2023.3335338}}

@ARTICLE{Real_World,
  author={Pawełczyk, Maciej Ł. and Wojtyra, Marek},
  journal={IEEE Access}, 
  title={Real World Object Detection Dataset for Quadcopter Unmanned Aerial Vehicle Detection}, 
  year={2020},
  volume={8},
  number={},
  pages={174394-174409},
  doi={10.1109/ACCESS.2020.3026192}}

@INPROCEEDINGS{MMAUD,
  author={Yuan, Shenghai and Yang, Yizhuo and Nguyen, Thien Hoang and Nguyen, Thien-Minh and Yang, Jianfei and Liu, Fen and Li, Jianping and Wang, Han and Xie, Lihua},
  booktitle={2024 IEEE International Conference on Robotics and Automation (ICRA)}, 
  title={MMAUD: A Comprehensive Multi-Modal Anti-UAV Dataset for Modern Miniature Drone Threats}, 
  year={2024},
  pages={2745-2751},
  doi={10.1109/ICRA57147.2024.10610957}
}

@article{RCSD-UAV,
title = {RCSD-UAV: An object detection dataset for unmanned aerial vehicles in realistic complex scenarios},
journal = {Engineering Applications of Artificial Intelligence},
volume = {151},
pages = {110748},
year = {2025},
issn = {0952-1976},
doi = {https://doi.org/10.1016/j.engappai.2025.110748},
url = {https://www.sciencedirect.com/science/article/pii/S0952197625007481},
author = {Wanxuan Geng and Junfan Yi and Ning Li and Chen Ji and Yu Cong and Liang Cheng}
}

@inproceedings{CST,
  title={CST Anti-UAV: A Thermal Infrared Benchmark for Tiny UAV Tracking in Complex Scenes},
  author={Xie, Bin and Zhang, Congxuan and Wang, Fagan and Liu, Peng and Lu, Feng and Chen, Zhen and Hu, Weiming},
  booktitle={Proceedings of the IEEE/CVF International Conference on Computer Vision},
  pages={6157--6166},
  year={2025}
}

@inproceedings{fasterrcnn1,
  title={Faster R-CNN: Towards Real-Time Object Detection with Region Proposal Networks},
  author={Ren, Shaoqing and He, Kaiming and Girshick, Ross and Sun, Jian},
  booktitle={Advances in Neural Information Processing Systems (NeurIPS)},
  year={2015}
}

@article{CMDet,
  title={Drone-based RGB-infrared cross-modality vehicle detection via uncertainty-aware learning},
  author={Sun, Yiming and Cao, Bing and Zhu, Pengfei and Hu, Qinghua},
  journal={IEEE Transactions on Circuits and Systems for Video Technology},
  volume={32},
  number={10},
  pages={6700--6713},
  year={2022},
  publisher={IEEE}
}

@misc{yolov5,
author = {Ultralytics},
title = {YOLOv5 by Ultralytics},
year = {2020},
howpublished = {\url{https://github.com/ultralytics/yolov5}},
doi = {10.5281/zenodo.3908559},
}

@INPROCEEDINGS{CascadeRCNN,
  author={Cai, Zhaowei and Vasconcelos, Nuno},
  booktitle={2018 IEEE/CVF Conference on Computer Vision and Pattern Recognition}, 
  title={Cascade R-CNN: Delving Into High Quality Object Detection}, 
  year={2018},
  volume={},
  number={},
  pages={6154-6162},
  doi={10.1109/CVPR.2018.00644},
  ISSN={2575-7075},
  month={June},}

@article{yolov3,
  title={Yolov3: An incremental improvement},
  author={Redmon, Joseph and Farhadi, Ali},
  journal={arXiv preprint arXiv:1804.02767},
  year={2018}
}

@article{cascaderpn,
  title={Cascade RPN: Delving into high-quality region proposal network with adaptive convolution},
  author={Vu, Thang and Jang, Hyunjun and Pham, Trung X and Yoo, Chang},
  journal={Advances in neural information processing systems},
  volume={32},
  year={2019}
}

@article{yolox,
  title={{YOLOX}: Exceeding YOLO Series in 2021},
  author={Ge, Zheng and Liu, Songtao and Wang, Feng and Li, Zeming and Sun, Jian},
  journal={arXiv preprint arXiv:2107.08430},
  year={2021}
}

@InProceedings{vitdet,
author="Li, Yanghao
and Mao, Hanzi
and Girshick, Ross
and He, Kaiming",
editor="Avidan, Shai
and Brostow, Gabriel
and Ciss{\'e}, Moustapha
and Farinella, Giovanni Maria
and Hassner, Tal",
title="Exploring Plain Vision Transformer Backbones for Object Detection",
booktitle="Computer Vision -- ECCV 2022",
year="2022",
publisher="Springer Nature Switzerland",
address="Cham",
pages="280--296",
isbn="978-3-031-20077-9"
}

@inproceedings{fd2net,
  title={Fd2-net: Frequency-driven feature decomposition network for infrared-visible object detection},
  author={Li, Ke and Wang, Di and Hu, Zhangyuan and Li, Shaofeng and Ni, Weiping and Zhao, Lin and Wang, Quan},
  booktitle={Proceedings of the AAAI Conference on Artificial Intelligence},
  volume={39},
  number={5},
  pages={4797--4805},
  year={2025}
}

@inproceedings{Xattn,
  author    = {Ashish Vaswani and Noam Shazeer and Niki Parmar and Jakob Uszkoreit and Llion Jones and Aidan Gomez and Łukasz Kaiser and Illia Polosukhin},
  title     = {Attention Is All You Need},
  booktitle = {Advances in Neural Information Processing Systems (NeurIPS)},
  year      = {2017},
  pages     = {599– – 609},
  url       = {https://arxiv.org/abs/1706.03762}
}

@INPROCEEDINGS{RTDETR,
  author={Zhao, Yian and Lv, Wenyu and Xu, Shangliang and Wei, Jinman and Wang, Guanzhong and Dang, Qingqing and Liu, Yi and Chen, Jie},
  booktitle={2024 IEEE/CVF Conference on Computer Vision and Pattern Recognition (CVPR)}, 
  title={DETRs Beat YOLOs on Real-time Object Detection}, 
  year={2024},
  volume={},
  number={},
  pages={16965-16974},
  doi={10.1109/CVPR52733.2024.01605}}

@article{yolov13,
  title={YOLOv13: Real-Time Object Detection with Hypergraph-Enhanced Adaptive Visual Perception},
  author={Lei, Mengqi and Li, Siqi and Wu, Yihong and Hu, Han and Zhou, You and Zheng, Xinhu and Ding, Guiguang and Du, Shaoyi and Wu, Zongze and Gao, Yue},
  journal={arXiv preprint arXiv:2506.17733},
  year={2025}
}

@inproceedings{TSFADet,
  title={Translation, scale and rotation: Cross-modal alignment meets RGB-infrared vehicle detection},
  author={Yuan, Maoxun and Wang, Yinyan and Wei, Xingxing},
  booktitle={European Conference on Computer Vision},
  pages={509--525},
  year={2022},
  organization={Springer}
}

@INPROCEEDINGS{OAFA,
  author={Chen, Chen and Qi, Jiahao and Liu, Xingyue and Bin, Kangcheng and Fu, Ruigang and Hu, Xikun and Zhong, Ping},
  booktitle={2024 IEEE/CVF Conference on Computer Vision and Pattern Recognition (CVPR)}, 
  title={Weakly Misalignment-Free Adaptive Feature Alignment for UAVs-Based Multimodal Object Detection}, 
  year={2024},
  volume={},
  number={},
  pages={26826-26835},
  doi={10.1109/CVPR52733.2024.02534}}
}


\end{document}